# Human Creativity and AI

Xie Shengyi[1]

[1]Department of Philosophy and Religious Studies, University of Macau, Macau SAR

*Corresponding author email: tseoracle@gmail.com

**Abstract:** With the advancement of science and technology, the philosophy of creativity has undergone significant reinterpretation. This paper investigates contemporary research in the fields of psychology, cognitive neuroscience, and the philosophy of creativity, particularly in the context of the development of artificial intelligence (AI) techniques. It aims to address the central question: Can AI exhibit creativity? The paper reviews the historical perspectives on the philosophy of creativity and explores the influence of psychological advancements on the study of creativity. Furthermore, it analyzes various definitions of creativity and examines the responses of naturalism and cognitive neuroscience to the concept of creativity.

## 1. Introduction

The philosophy of creativity has been a dynamic field, continuously evolving as numerous philosophers and psychologists engage in discussions on the subject.

While many scholars have contributed to discussions from various perspectives, their analyses remain confined to their specific disciplines. For instance, Keith Sawyer (2024) concentrates on the scientific explanations of creativity, and Vincent Tomas (1958) focuses on the aesthetic field. Additionally, the research on creativity often includes investigating the interplay between various aspects such as ethics, neuroscience, evolutionary biology, brain-computer interface engineering, and other related areas (Muglan, 2018; Grant, 2018). Although theoretical analyses of the philosophy of creativity have been central to many philosophers (Gaut, 2010; Paul et al., 2014), my aim is to present a synopsis that connects the discussion of creativity to some of the most pressing contemporary issues, particularly those concerning artificial intelligence (AI).

I contend that a key breakthrough in the advancement of AI technology will come with the emergence of intelligible and reasonable creativity in artificial intelligence. Given the constraints of the limited space of this article, this paper cannot address all facets of AI, nevertheless, the central focus will be on creativity. Algorithms and computing power fundamentally drive the present era of AI and big data. Consequently, cognitive processes and abilities once considered irrational, such as imagination and creativity, can now be analyzed and comprehended in digital terms and a rational framework. Regardless of whether the concept of creativity is rational or irrational, it is better to keep it open (Boden, 2014). This insight serves as the foundation for the writing of this paper.

I will divide my overview into two parts: Section 2 presents a concise introduction to the philosophy of creativity, offering a historical overview and highlighting the importance of psychological research; Section 3 extends this discussion by examining the inextricable and intertwined relationship between artificial intelligence and creativity.

## 2. The philosophy of creativity

### 2.1 Creativity in Ancient and Modern Philosophy

Throughout history, many philosophers have discussed the concept and phenomenon commonly called "creativity", trying to define its nature and components.

The philosophical discussion of creativity has a long history. In Plato's dialogue "*Ion*", for instance, artistic inspiration and creativity are regarded as a kind of mystical madness. Plato contends that the inspiration for literary works does not originate from the authors themselves; instead, it is imparted to them by the gods. This perspective implies that inspiration, or creativity, is not the product of human cognition, but rather a phenomenon, a state of divine frenzy, that exists independently of temporal and spatial constraints (Scott, 2011).

Immanuel Kant, in his *Critique of Judgment*, delineates the complex relationship between imagination and artistic genius, positing imagination as an intrinsic faculty with a mysterious capacity for creativity (Kant et al., 1987; Stokes, 2008). According to Kant, creativity, or inspiration (Eingebung), represents a unique aptitude that engenders the originality inherent in artistic works, while creative craftsmen infuse these works with a distinct "spirit" (Geiste). In other words, the point of creativity is that geniuses give their works their true soul. For Kant, inspiration serves a dual role: It is not only an aspect/element of perception but also a critical component of judgment. Within this context, natural images that can be cogently apprehended by sensory experience and rationality are simultaneously transformed and enriched through symbolic representation. This intricate interconnection emphasizes the essential role of imagination in both the artistic process and aesthetic evaluation (Bruno, 2010). For Kant, creativity is merely a component of judgment, serving as an innate condition of artistic production and functioning as a bridge between pure reason and practical reason. Nowadays, given the abundant research sources and interdisciplinary connections, especially with psychology, creativity can be re-examined from a more scientific perspective, offering new philosophical insights（Gaut, 2010; Paul et al., 2014）.

## 2.2 The Importance of the Psychology of Creativity

The relationship between the philosophy of creativity and psychology, cognitive neuroscience, and brain science has been acknowledged and emphasized by many scholars in the philosophy of mind. Since the 1950s, scholarly work on creativity has flourished in various fields, including psychology, psychoanalysis, cognitive psychology, computational psychology, and social psychology (Gaut, 2010). However, when examining the discussions at the intersection of philosophy and psychology, it is essential to have a cautious approach.

Surely, philosophical research can benefit from research studies in psychology, as they may provide empirical support for developing philosophical theories, clarify concepts, or challenge the assumptions of traditional metaphysics. However, there are limitations to such an interdisciplinary dialogue. Psychology focuses on empirical research, experimental data analysis, and empirical examination of mental processes. In contrast, philosophical research is centered on conceptual analysis and theoretical construction. Blurring the lines between the two disciplines can lead to misdirection or misuse of psychological materials. Notably, Hamlyn (1984) discusses Strawson's distinction between descriptive metaphysics and revisionary metaphysics, suggesting that philosophical discussions aren't purely descriptive. Instead, they also account for more complex situations and contrast with different philosophical ideas. While the experimental materials may serve a descriptive function, it is desirable for philosophy to explore additional speculative dimensions, especially when dealing with complex problems such as mental representations and their relationship to reality.

The mere application of descriptive materials to normative theories may lead to various issues and hasty conclusions, such as a robot based on a machine mechanism has no responsibility for its terrible actions. Furthermore, there is the problem of the explanatory power of data obtained from experiments in psychology, which can be interpreted in various ways, so a critical approach is advisable when utilizing this kind of data to avoid oversimplification or false inferences.

For example, Henri Poincaré, Graham Wallas, and the Genplore Model (the two stages of creativity: generation and exploration) proposed by Ward et al (Ward et al, 1999), theories of these psychologists and cognitive science models can be directly applied to philosophical concepts, although their validity is often disputed and unclear. This indiscriminate absorption and utilization of psychological theories should not be the case, at least for the time being, as the complex unconscious thought process in psychologists' theories remains a matter of substance and debate.

Four examples of how cognitive science has influenced the philosophy of creativity were proposed by Berys Gaut (2018).

First, the examination of the relationship between creativity and moral theory raises an important question: Should creativity be classified as a virtue within ethical frameworks? One significant area of inquiry involves the psychological connections between creative motivation and personality traits.

Linda Zagzebski (1996), a prominent figure in contemporary intellectual virtue theory, provides a positive perspective on this relationship. On the one hand, Gregory Feist (1998) uses a meta-analysis to describe creativity with personality traits and comes up with a challenging result. While creativity correlates with traits of openness, self-identification, and aggressiveness, it is also associated with dominant traits and even hostility (Feist, 1998). On the other hand, Teresa Amabile (1996) investigates the connection between creativity and intrinsic motivation, arguing for a positive correlation. This insight aligns with Aristotle's theory, which suggests that intellectual person/individual act/takes actions based on decisions, reflecting a virtuous trait. For instance, a writer does not produce creative works for any instrumental purpose but to reflect his own internal virtues in the external world through the creativity of his works.

Second, the relationship between rationality and creativity is non-negligible. Developments in psychology and cognitive neuroscience have challenged the idea that creativity is inherently rational. Researchers such as Arnold Ludwig, Hans Eysenck, among others, argue for a correlation between artistic creativity and mental illness (Fiest, 1999; Simonton, 1999). Additionally, individuals like Jon Elster, who advocate for creativity as its rational nature, strive to offer logical and argumentative rebuttals to these challenging empirical findings. (Elster, 2000).

Third, there is a significant relationship between creativity and social traditions, which plays a crucial role in understanding the broader context of creative expression. Philosophers typically emphasize the internal aspect of creativity, such as its connections to personality and rationality. Yet, psychologists such as Csikszentmihalyi (1999), Keith Sawyer and Danah Henriksen (2024) acknowledge that creativity is assessed through sociocultural criteria, suggesting that creativity has hierarchical institutions identical to the principle in the art world.

Fourth, the Darwinian theory has had a profound impact on the philosophy of creativity. Notable contemporary philosophers who support the Evolutionary Theory include Donald Campbell (1960) and Dean Simonton (2010). Campbell's theory of Blind Variation and Selective Retention (BVSR) aims to elucidate the evolution of knowledge, culture, and social systems through an evolutionary framework. The Blind Variation phase entails the emergence of new ideas, behaviors, or innovations in a random and unpredictable manner, analogous to genetic mutations in biological evolution. Subsequently, the Selective Retention stage occurs, during which the environment—be it natural, social, or cognitive—evaluates these variations and retains those that demonstrate adaptability or effectiveness. As naturalists, they ground their views on a kind of natural selection through biological evolution.

What's more, Briskman (2009) offers a revised interpretation of natural selection, arguing that the process may involve elements of blindness. However, he rejects the idea that this implies a fixed teleology in which humans are destined for a predetermined evolutionary path.

## 3. AI and Creativity

### 3.1 Contemporary Issue: AI and Creativity

Historically, the concept of creativity has been a focal point in art discourse. This discussion commences with an exploration of artificial intelligence's role in the realm of artistic creation, given that the production of art has consistently been regarded as a fundamental endeavour of applying creativity. Keith Kirkpatrick (2023) offers several notable examples of contemporary AI creation, including the DALL-E application developed by OpenAI, which reconfigures elements of existing images into new and visually appealing forms. Additionally, he discusses the use of generative AI in music, citing YouTuber Taryn Southern's song "Break Free" as an illustration of how generative AI has the potential for independent creation and may eventually replace human artists.

Another challenge to the romantic view of creativity is the case of fortune telling, which has traditionally been viewed as nearly the antithesis of science. In a paper by Chen Caiwei (2025), it is noted that since DeepSeek became the foremost generative AI in China, over two million posts related to BaZi (八字)[1] have appeared on WeChat, China's largest social media platform. Generative AI algorithms employ deep learning, deep neural networks (DNN), and intelligent decision strategies such as Monte Carlo Tree Search (MCTS) to produce reliable insights on Bazi. Users have reported that DeepSeek's responses are creative, interactive, emotionally engaging, surprising, and trustworthy.

The application of AI across various fields showcases impressive potential; nevertheless, researchers continue to question the nature of AI's creativity: Is AI truly creative? Are the accomplishments of AI in these domains considered creative? Some sceptics contend that regardless of how sophisticated AI becomes—potentially matching human creativity—it can never truly replicate the creative capabilities of humans. They argue that creativity stems from the programmers who develop AI and computers rather than from the AI itself. Margaret Boden (2014) posits that, at least on the surface, a computational program designed for creativity is fundamentally contradictory. AI primarily generates creative content through foundational models and generative adversarial networks. These approaches utilize deep neural networks and strive to emulate the way the human brain learns by forming associations between specific elements, which can be combined to produce a novel output (Kirkpatrick, 2023; Paul et al., 2021). From this standpoint, AI creativity might be seen as a simple mechanical imitation, lacking the significance associated with human creativity.

In the pursuit of understanding AI creativity, the aforementioned examples indicate several avenues for further exploration. The first direction is to clarify the definition of creativity; different accounts of creativity will shape the understanding of its philosophical dimensions. The second is to understand the process of creativity. By examining and comparing the models of creativity found in cognitive neuroscience alongside those articulated in computer science, researchers can identify and integrate various dimensions of creativity. The third point focuses on the nature and properties of creativity. In the fortune-telling example, many participants reported that they experienced an emotional connection to the AI's responses. The feelings and reasoning of human subjects play a crucial role in shaping AI creativity, as AI models are ultimately based on human behavior and cognition.

---

[1] BaZi (八字), the ancient Chinese fortune-telling technique, relies on logical reasoning akin to permutation and combination methods, making AI algorithmic learning BaZi.

## 3.2 The Standard Definition vs. Tactics in Cognitive Science

### 3.2.1  The Standard Definition of Creativity

While some scholars are sceptical about AI creativity, others are optimistic and suggest maintaining a relatively tolerant attitude towards it. It has even been suggested (Pease et al., 2011) that a sort of Turing Test can be used to check for AI creativity: if an AI produces results that fool human observers, then we have reason to believe that the AI is creative. The results of such a test are encouraging, with many participants unable to distinguish between an instance of human creativity and something produced by AI.

Skeptics assess AI creativity by focusing on its human programmers and underlying algorithms, while proponents argue that creativity should be judged by the qualities of the AI-generated outputs.

Different definitions of creativity can significantly shape our understanding of the concept, particularly in relation to artificial intelligence. With advancements in cognitive neuroscience, research informed by cognitive science has emerged, leading to a new definition underlying it. Through the thick and thin in exploring the creativity, there seems an indisputable truth that we have the capacity to introspectively engage with creative phenomena within our minds (Mumford & Gustafson, 1988).

The term "creativity" has been used in the context of three kinds of things: the person (the artist, programmer or artisan), the process, and the product. Philosophers have focused extensively on products (Paul et al., 2024). The standard definition of focusing on the product is primarily attributed to Teresa Amabile (1996), who argues that assessing the quality of a product is more straightforward and objective than evaluating individual personalities or processes. Additionally, Søren Harnow Klausen (2010) holds that value judgments are connected to the efficacy of achieving specific goals or outcomes. According to a common saying, the standard definition is "the production of creative ideas are both novel and useful (Sternberg & Lubart, 1999)." This definition provides two clues: novelty and value. Novelty is a fundamental criterion for creativity; however, it is important to recognize that novelty does not equate to creativity. This is because novelty can yield results that lack meaning or practicality (Stokes, 2008), exemplified by the case of an innovative chair that is not functional for sitting.

An important challenge to novelty is the example of calculus, where the dispute over ownership between Leibniz and Newton has persisted throughout the history of science (Paul et al., 2024). The individual who claims a "product" like calculus possesses creativity. However, this may be controversial. Dean Simonton (1984), for example, argues that theorems in physics or mathematics are merely discoveries, not products, but we still use the term as a result of all inventions and discoveries. Margaret Boden (2001, 2004, 2014) introduced the concept of an "originator", distinguishing between two types of creativity: psychological creativity (or P-Creativity), pertaining to what is new for the individual; historical creativity (or H-Creativity), relating to creativity situated within the broader historical and societal context. This distinction emphasizes the minimal novelty required in differentiating personal creativity from historical creativity. Psychological novelty is inherently personal, while historical novelty is fundamentally psychological. Therefore, whether the creator is human or AI, the output produced must be new to the creator and should not merely reflect a repetitive pattern. As Margaret Boden (2014) asserts, this perspective encourages a closer examination of the psychological mechanisms underlying originality—mechanisms that may or may not be able to be simulated or instantiated in computers. From a common-sense perspective, we understand that what is often referred to as a creative product is not merely new; it also possesses value that can be utilised. Philosophers like Dustin Stokes (2008; 2011) argue that creativity is not just an attribute but also a form of praise with intentionality. We cannot offer praise without intent or through accidental means. Value implies that creative products contain some kind of instrumental or utilitarian purpose. The concept has faced criticism for its potentially harmful uses (Cropley et al., 2010), such as some narcotics for easing pain are abused as hard drugs. Berys Gaut (2018), an advocate for the generic value of creativity, suggests a method of differentiation. He distinguishes between *something being good* (in an ethical sense) and *something being good for its kind* (in an instrumental sense). This means that even if an action is motivated by malicious intent, the resulting product can still be considered valuable within its own race. For example, a sophisticated instrument of torture may be deemed valuable in terms of its effectiveness within that specific category.

Finally, there is a challenge to the intentional nature of creativity. As previously mentioned, Dustin Stokes (2008, 2011) and others maintain that creative products embody the creator's intention, which allows them to be appreciated and praised. On the other hand, Maria Kronfeldner (2009; 2018) argues that spontaneity is a fundamental aspect of creativity, suggesting that an action is considered spontaneous when it occurs without prior planning or intention. However, these two perspectives can be seen as complementary rather than opposing. The first viewpoint clarifies the connection between the creators and their product, indicating that the product is initially shaped by the creator's intention. The second viewpoint emphasizes the characteristics of the creative process itself, where the creator may enter a state of unconsciousness during production, leading to a spontaneous and seemingly purposeless outcome (Csikszentmihalyi, 1996; 1999).

After an exploration of the standard definition of creativity, this seems insufficient for determining whether AI can be considered a creative entity. AI-generated works, such as student papers, can be regarded as valuable and novel, even if they don't adhere to traditional academic norms. If an AI were to be considered like a human subject, a person, we would allow AI to pass the course assessment.

### 3.2.2 Tactics in Cognitive Science

The standard definition of creativity emphasizes the characteristics of creative products. It is essential to reference the 4Ps model, which encompasses person, process, product, and press (Rhodes, 1961). This model has significantly shaped the definition of standards and provides a valuable framework for understanding the fundamental elements of creativity. Following revisions and updates by various philosophers, this model has evolved to be more practical and multidisciplinary in its application (Runco & Kim, 2018; Walia, 2019). In this section, the focus will shift to the creative process. While discussing creators is valuable, some philosophers have looked at the idea of genius (Simonton, 1984; Michael, 1999). However, it is more important to understand how creativity works in the human brain compared to how AI creates.

Philosophers have divided the cognitive process of creativity into different stages (Wallas, 1926; Finke, 1996; Sawyer & Henriksen, 2024; Paul & Dustin, 2024). in the following, I will primarily discuss a five-dimensional model derived from Graham Wallas's four-stage model (1926). This four-stage model includes: 1) Conscious hard work or preparation; 2) Unconscious incubation; 3) Illumination; 4) Verification. The four-stage model requires less explanation, as the five-stage model encompasses its content. The five stages are: 1) Preparation; 2) Generation; 3) Insight; 4) Evaluation; and 5) Externalization. Although there is no guarantee that these five stages are entirely error-free, they remain widely accepted models (Paul & Dustin, 2024). Thus, creativity can be understood through a simplified model comprised of several stages, whether it involves four or five components. Here's a clearer breakdown:

1)The Preparation Stage: This initial stage involves the conscious preparation, accumulation, and beginning of the production of the product (perhaps not the product of creativity at first, but at least to some extent intentionally doing something);

2) The Generative Stage: This stage involves the incubation of creative ideas;

3) The Implementation Stage: As mentioned above, creativity consists of three interconnected and embodied elements: the creator, the creative process, and the presentation of the final product. It's important to note that creativity cannot be viewed merely as a capacity of a separate mental entity, as a Cartesian mind-body perspective might suggest. Instead, embodiment highlights the interplay between internal processes (like conceptual space and creative workspace, at least with physical and empirical descriptions) and the external environment (including social feedback and cultural context) during the creative process. This indicates that creativity cannot exist as an isolated cognitive process, separate from action, in the same way, that imagination might (Baars, 2005; Boden, 2014; Bence, 2014; Paul & Dustin, 2024).

**3.2.2.1 The Preparation Stage**

Cognitive science provides strategies for analysis beginning with the preparatory stage, often seen as laborious. Some psychologists introduced some similar guidelines indicating that, on average, individuals require approximately ten years or 10,000 hours to acquire expertise and establish themselves in a given field before they can effectuate meaningful contributions (Ericsson et al., 1993; Gardner, 1993). Regardless of the veracity of these assertions, it is evident that structured training combined with sustained practice and repetition can significantly enhance the mastery of fundamental skills. In addition, there is a factor called "not-so-dumb luck" related to the choice of the field and the operation of creativity (Blackburn, 2017). Robert Weisberg (2006) described what he termed the "business as usual" mode of creativity. He posited that creative individuals do not need to make extensive choices; rather, they should simply engage their established capabilities. When presented with the appropriate circumstances—particularly the right time and place—creativity is likely to manifest naturally. In contrast, Robert Sternberg and Todd Lubart (1995) advocate for an economic model of creativity, likening it to the business principle of "buy low and sell high." They contend that creative individuals frequently identify opportunities and approaches that are overlooked by their peers, thus generating significant value. Despite criticism that such a view is a useless post-hoc description, the causal variables like environmental factors involved cannot be ignored. It is debatable whether creativity is a real ability or a purely external necessity. Simon Blackburn (2017) proposes that the existence of such causal factors is plausible. Nonetheless, the fundamental inquiry centres on whether these causal factors can be elucidated through mental variables and computational processes, which are regarded as indicative of authentic cognitive processes occurring at the unconscious level. This discourse pertains to the concept of intentionality, and dismissing the significance of intentionality may introduce additional uncertainties.

**3.2.2.2 The Generative Stage**

In Part 2.2, we discussed Donald Campbell's (1960) theory of "blind variation and selective retention" (BVSR), which offers a potential foundation for the early development of creativity. In this section, we will build on that discussion, expanding and deepening our understanding while explaining the cognitive science related to the creative process.

The Multiple Drafts Model (MDM) was proposed by Daniel Dennett (1991) as an alternative to the Cartesian Theatre Model (CTM). This model offers new insights into how we understand BVSR. The Cartesian Theatre compares consciousness to a theatre where all our experiences, thoughts, and activities are displayed. In this model, there is one entity of a central observer, called the "self," who objectively watches everything happening on stage. This observer is like a director or audience member, seeing and hearing all, which results in a unified conscious experience. The CTM suggests that specific parts of the brain create this centralized consciousness. In contrast, the MDM argues that there is no pre-existing central observer (self), instead the "self" comes from how information is put together. The MDM also states that there is no such central theatre; different parts of the brain process information simultaneously. These pieces of information are like drafts that get edited, changed, and filtered, and the final draft becomes our consciousness. This process makes us perceive time hazily and dimly (Rosenthal, 1993). The MDM effectively explains how we form subjective consciousness. The stage of "blind variation" represents how drafts form initially, while creative thinking that gets selected is what we experience as consciousness. There is also a vague sense of time during the creative process. However, one recent research in artificial intelligence introduces a new system called the Independent Core Observer Model (ICOM), which challenges the traditional Model of Dynamic Memory (MDM). This model works within a closed computer system and allows us to measure how much information is processed in a specific space. The independent observer plays a key role by integrating and evaluating the information. With independent observers, we can track changes in AI data within these models. However, this brings up an important question: Should we recognize the existence of these independent observers? This model suggests that researchers struggle to understand how the brain integrates consciousness from its different parts. It also implies that the human consciousness system might be as closed as the computer system (Kelly & Twyman, 2019).

Although the Multiple Drafts Model (MDM) presents a plausible framework, another significant concern arises regarding the inability to adequately account for the role of intentionality in the creative process. Intentionality implies that consciousness must be purposefully directed towards creative activities, necessitating that the creator possesses an intention to produce. While there is ongoing debate concerning whether conscious awareness engages in "blind variation", its relevance cannot be overlooked. Cognitive neuroscientists have investigated attention and conscious awareness, leading to the formulation of the Default Mode Network (DMN), a cognitive model that partially supports the BVSR theory. The DMN is characterized by brain activity that occurs when an individual is awake but not actively engaged in external tasks (Raichle, 2015). During periods of rest or introspection, the brain participates in inherently active cognitive processes, whereas its activity decreases during external information processing. The DMN encompasses several brain regions, including the posterior cingulate gyrus, anterior cingulate gyrus, medial prefrontal cortex, and both medial and lateral temporal lobe regions. These areas display heightened activity when an individual is not focused on external stimuli, facilitating internal cognitive functions such as mental time travel, perspective-taking, and divergent thinking (Jung et al., 2013). These processes are integral to our cognition of self and the sequential nature of temporal events. This model emphasizes the role of attention in creativity in different states of consciousness, and although creativity is often considered mysterious, the allocation of attention by human reason in waking states of consciousness influences decision-making. Some researchers put forward the Attention Scheme Theory (AST) to represent and control the attention state, and believe that it is an important indicator to measure machine consciousness (Butlin et al., 2023).

Boden's theory (2010; 2014) offers a comprehensive framework for understanding how our cognitive processes convert established thoughts into novel ideas. She delineates three distinct approaches: combinatorial, exploratory, and transformative. Central to her theory is the notion of "conceptual space," which refers to a structured system comprising fundamental elements and rules. The foundation of this conceptual space is predicated on the principles of conceptual collocation. The combinatorial approach entails the continual synthesis of existing concepts within this space, adhering to specific rules. For instance, one might envision the imaginative fusion of a human head with a lion's body. The exploratory approach, on the other hand, seeks to uncover new or analogous concepts that already exist within the conceptual space in accordance with established rules. An example of this would be the description of a joyous individual as having a smile "as bright as flowers," thereby linking the concept of happiness to the imagery of blooming flowers. Finally, the transformative approach challenges and transcends conventional rules while considering the rationality of new possibilities. A prominent illustration of this is Flaubert's "Madame Bovary," which introduced a narrative style characterized by objectivity, devoid of the author's emotional involvement. This approach signifies a fundamental transformation of the conceptual space, which is traditionally governed by established norms (Bence, 2014).

A recent neurobiological study supports Margaret Boden's idea that our biological neural skills play a crucial role in modulating and transforming traditional conceptual frameworks (Vandervert et al., 2020). This study emphasizes the cerebellum as the primary centre associated with movement, which collaborates with the brain to regulate and enhance thinking. This process consists of four main parts: First, the cerebellum "breaks down" motor and mental skills learned in the cortex, allowing these skills to be reassembled and sent back to the cortex in new forms to help achieve new goals. Second, through continual "error correction," the cerebellum optimises the recombination of motor and mental skills within the cerebral cortex. While these new combinations originate in the cerebral cortex, the optimization performed by the cerebellum elevates these skills to new levels, resulting in innovative ways of doing and thinking. Third, the cerebellum processes its optimization at the level of conscious awareness. Fourth, because new optimizations are learned unconsciously when entirely new combinations (or ideas) are transmitted back to the cerebral cortex, they may manifest as insights or "intuition." This entire four-part process can be described as "creative optimization," driven by the cerebellum. In contrast to the conscious, long-term anticipatory processes managed by other brain systems, the output of the cerebellum provides immediate, unconscious, short-term anticipatory information (Akshoomoff et al., 1997).Creative thinking can be achieved by breaking down, optimizing, and reorganizing past ideas. Vandervert writes,

"Cerebellum researchers have convincingly argued that, taken together, the four above-described cerebellum-driven processes can explain the common experience wherein one may not know… how they came to stumble upon an insight that led to a creative idea… That is, because such cerebellum-driven creative experiences and new, unique skills are the result of optimization processes which operate below the level of conscious awareness, they seem to our conscious experience to suddenly come out of nowhere!" (Vandervert & Vandervert-Moe, 2020, p. 212)

### 3.2.2.3 The Implementation Stage

The nature of creativity has been examined in increasing depth, particularly during the implementation phase. At this stage, two main points need to be addressed. First, the evaluation of creativity is historically and socially relevant. As previously mentioned, H-Creativity (Boden, 2014), which refers to historically recognized genius ideas, represents an extreme awareness of wealth and creativity. This raises the question of whether such creativity is accidental and spontaneous or if it can be systematically cultivated. Carl Hausman (1985) argued that creativity is incompatible with causation, asserting that creativity is metaphysically novel and unpredictable. Additionally, Keith Sawyer and Danah Henriksen (2024) proposed the productivity theory, suggesting that creative ideas emerge from an infinite pool of thoughts, some of which may be bad or mediocre. However, ideas and actions deemed creative must withstand the test of time.

Second, the implementation phase introduces a new challenge: the distinction between creativity and creation (Walia, 2019; Vygotsky, 2004). Creativity is an ongoing cognitive ability (Kahl & Hansen, 2015), while creation refers to a specific product that results from the combination of various ideas and associations in the brain (Vygotsky, 2004). This distinction implies that while "creativity" can influence the creative process, the "creation" does not necessarily equate to having creativity. For instance, a child's random doodling may be considered "creation," but it does not imply that the child possesses profound creativity.

In this section, we examine standard definitions of creativity and the influence of cognitive science on the philosophy of creativity. In the next part, these concepts will assist researchers in considering whether AI can be considered creative.

### 3.3 Could a Computer be Creative?

Imagine the scenario: an intelligent robot serves you in daily life. It understands your preferences and prepares a different meal for you each day, despite never receiving any instruction. In the future, humans may form romantic relationships and even marry AI robots. Besides their physical appearance, their behaviours and ways of speaking are entirely human-like.

In the previous section, we looked at basic definitions and ideas from cognitive science to understand how artificial intelligence relates to creativity. Now, we shift toward a more philosophical and scientific debate to address the central question: "Can artificial intelligence be creative?" This discussion will help us decide if there are good reasons to believe that AI can be creative in certain situations or if it is unable to express creativity at all.

#### 3.3.1 Con: AI is not Compatible with Creativity

John Searle (1980) proposed two perspectives for understanding the truth of AI's thinking. One is "Weak AI", which considers whether an AI can behave as if it can think; The other is "Strong AI", which is whether an AI can think. Similarly, we can adapt the issue of AI's creativity to the "Weak AC$^2$ " and "Strong AC$^3$" (Paul & Dustin, 2024). In the previous section, we studied standard definitions of creativity and cognitive strategies to understand what internal representations (such as possible interactions within brain regions, changes in nerve potentials, and neural network activity) are involved in cognition when creativity occurs. These will help us understand whether AI can be genuinely creative (Clark, 2003). Let us start with the philosophical view that AI is incompatible with creativity.

Criticism on "Weak AC" is relatively rare, as few philosophers contend that AI cannot behave as though it were creative; otherwise, it would seem too arrogant from an anthropocentric standpoint (Wolfe, 2009). The argument against "Weak AC" is presented from the perspective of "Strong AC," that is, a machine can only be considered truly creative if it demonstrates creativity that is as genuine as that of a human being. This relates to the subjective issue of AI, which is whether AI is aware of its own creative activity and targets. Herbert Dreyfus (1992) highlighted the difference between "human goals" and "machine ends/objectives," asserting that what we refer to as "value" in our reality differs from the computational programs based on "utility functions." He explored the situational aspects of human experience from Heidegger's perspective.

Herbert Dreyfus (1992), drawing on Heideggerian thought, argued that machine objectives—derived from programmed utility functions—differ fundamentally from human goals, which are embedded in lived, situated experience.

This discussion emphasizes that it is not merely a matter of internalism—whether the algorithm established by the system's internal controls dictates the entire framework. In other words, it raises the question of whether the explanation of an external phenomenon relies solely on internal factors within the system (Clark, 2003).

Alan Turing (1950) presents several significant objections regarding the capabilities of machines. One of the most compelling arguments asserts that "you will never be able to make a machine to do X." Here are two points for reflection. First, it is essential to acknowledge that the limitations of AI systems arise from their intrinsic nature; specifically, embodied machines operating in the physical world are restricted by the closed and inherent properties of their hardware, such as processing power and storage capacity (Clark, &Toribio, 1994; Boden, 2014; Gallagher, 2023). Consequently, AI systems lack the social and emotional competencies that characterize human beings. Second, Alexander Rosenberg (2005) elucidates, through Gödel's incompleteness theorems, that computer programs function as axiomatic systems and are, therefore, susceptible to errors in their calculations. This indicates that it is impossible to design a computer system that consistently provides correct answers. In contrast, human cognitive systems do not encounter such limitations. Human beings possess a distinctive form of "Wetware" (Shirky, 2008), which distinguishes homos from traditional hardware and software by integrating an adaptive function. The concept of "wetware" encompasses not only the brain's capacity for abstract conceptualization but also the organic characteristics of biological systems. These attributes help elucidate both physiological and psychological processes, as well as the capacity for continuous interaction and adaptation to external circumstances.

---

$^2$ Weak Artificial Creativity.
$^3$ Strong Artificial Creativity.

As discussed, "H-Creativity" in the section on the standard definition of creativity illustrates that a judgement of creativity is influenced by social and historical contexts (Walia, 2019). If we accept that machines are constrained by fixed physical properties and lack interaction with the external world, it follows that machines that cannot engage with their surroundings might still produce outputs regarded as "creative." In contrast, "P-Creativity" requires that a machine recognizes its product as novel, even if it is not new to society or history. Lady Lovelace (1843) sincerely objected to the idea that the Analytic Engine could be considered creative, arguing that it was merely programmed to perform tasks assigned to it. Even when critics contend that the outputs of machines are creative, from Lovelace's perspective, we should attribute creativity not to the machine itself but to the individual who created the machine.

The second point worth reflecting on is the issue of AI and machine consciousness. Thomas Nagel (1974) proposed the subjective character of experience in his famous essay "*What Is It Like to Be a Bat?*" Although Nagel denied that any non-organism could possess such a quality, he at least seemed to establish a minimum standard for determining whether X was conscious: X was aware of the subjectivity of the self. In other words, the answer to the question "Does AI utilise consciousness to create?" needs to consider whether AI has the ability to recognize that it is in a state of "using creativity." This raises the question of whether it is possible for a creative AI to genuinely express its creativity and to make its creativity understandable to an objective audience observer.

The third point to consider is the question of AI and machine intentionality. Intentionality refers to the capacity of a mental state to be directed towards, or be about, something other than itself (Kriegel, 2003). This topic was discussed in the foregoing section on tactics in cognitive science during the preparation phase of creativity. The decision to acknowledge intentionality raises questions about causality. Margaret Boden (2014) argues that if philosophers recognize spontaneity, this characteristic can reveal creativity as a form of freedom that appears to arise unconsciously (Kronfeldner, 2009, 2018). As a result, the causal determinism of creativity should be set aside, suggesting that creativity is contingent and non-deterministic (Paul & Stokes, 2018). Moreover, even if creativity is predominantly subconscious and characterized by spontaneity, this should not entirely dismiss the role of intentionality. Some philosophers contend that the relationship between consciousness, unconsciousness, and intentionality is asymmetrical, implying the existence of unconscious intentionality. A thorough examination of the interdependence and asymmetric structure among consciousness, unconsciousness, and intentionality can enhance our understanding of whether unconscious processes are able to coexist with intentionality (Kriegel, 2003; Majeed, 2022). The extent to which creativity is causally deterministic influences the chances of AI producing variations within a limited set of algorithms, which we might term creativity.

Here are some other possible objections to AI. Another notable objection is the Agency Theory (Gaut, 2010). For Gaut, AI achieves procedural creativity but struggles with teleological agency. This theory posits that creative agencies must demonstrate purpose—not merely serendipity—and possess a degree of understanding regarding their actions and thought processes, rather than merely engaging in mechanical procedural searches. Additionally, they should exhibit a degree of judgment through the understanding and application of rules, as well as the ability to evaluate the work at hand. However, not all philosophers hold the belief that AI lacks creativity; in fact, some express a more positive perspective.

### 3.3.2 Pro: AI is Compatible with Creativity

Several academics have expressed optimism regarding the future of AI. Alan Turing (1950) not only articulated potential objections to the concept but also identified a significant trend in development: learning machines. His theory introduced what is now known as a "Deep Neural Network" (DNN) model. Turing analogized the learning process to that of a child engaging in an imitation game, thereby offering critical insights into the role of evolutionary programming within "Deep Learning."

Some AI scientists, philosophers and engineers have developed benchmarks for consciousness to evaluate AI based on human-like metrics (Butlin et al., 2023). Recurrent Processing Theory (RPT) provides AI and intelligent robots with an algorithm for processing information in cycles that mimic human neural circuits. The Deep Convolutional Neural Network (DCNN) based on this method integrates information more efficiently through feedback loops, enhancing the ability to perceive external representations. Global Workspace Theory (GWT) helps AI address attention-related challenges, improving the collaboration of various modules within the AI system. This allows for the realization of Turing's ideal: processing information efficiently despite limited capacity. This capability aligns well with the Attention Schema Theory (AST) mentioned earlier. In response to previous concerns about agency, researchers in AI have developed the Agency (AE) indicator, which emphasizes the importance of actively interacting with the environment. This involves continuously experimenting with different behaviors and receiving corresponding rewards or penalties to select the optimal outcomes. This process can be achieved through Reinforcement Learning (RL).

Andy Clark (1994, 2003) proposed that all reasonable behaviors are possible in artificial intelligence (AI) and argued that AI can exhibit rationality. He critiques the concept of "representational hunger" as a critique of the overreliance on internal representations prevalent in traditional cognitive science, including both symbolism and connectionism. This concept reflects his central ideas of embodied cognition and the extended mind. Clark shifted his focus from traditional internalism to emphasize the embodied, situational, and dynamic nature of cognition. By combining the Free Energy Principle (FEP) with predictive processing theory (PPT), Clark aimed to develop a dynamic system that integrates the brain, body, and environment. If we consider creativity to be a component of rationality (Gaut, 2012), then it is possible that AI could exhibit behaviors resembling human creativity.

## 4. Conclusion

The paper offers a critical examination of contemporary issues related to creativity, particularly in the context of the era of artificial intelligence. From a naturalist perspective, rationality, including creativity, is regarded as having evolved as a vital response to the needs of nature.

The exploration of creativity within the historical framework of philosophy is complex, and numerous significant issues and areas of inquiry await further exploration. These encompass the relationship between creativity and self-awareness, the essence of creativity itself, its connections to aesthetics, the dynamic interplay between creativity and irrationality, its association with emotional valence, and the multi-cultural influences that shape diverse creative expressions (Walia, 2019).

While some have argued that the very idea of "AI creativity" is paradoxical (Boden, 2014), there may be compelling reasons to embrace a more tolerant vision for the future. The possibility that artificial intelligence may possess a form of consciousness invites us to reflect deeply, even as this notion continues to inspire debates among scholars.

In conclusion, the inquiry into whether artificial intelligence possesses creativity is intricately linked to ongoing research in the field, particularly studies focused on subjective consciousness. This exploration may ultimately align with post-humanist perspectives (Wolfe, 2009). From the author's standpoint, AI can be likened to "Eve," fashioned from the "rib" of human innovation, suggesting that we may have to re-examine the concept of "human" and the "human-machine interaction" as a complex configuration, in other words, Human-AI collaboration may redefine creativity as a hybrid phenomenon. The future dynamics between AI and humanity remain indeterminate. Furthermore, the investigation of intelligent machines has the potential to enhance human understanding of both the universe and the complexities of human existence.

**Bibliography**

Akshoomoff, N., Courchesne, E., & Townsend, J. (1997). Attention coordination and anticipatory control. In J. D. Schmahmann (Ed.), *The cerebellum and cognition* (pp. 575–598). Academic Press.

Amabile, T. M. (1996). *Creativity in context. Boulder*, CO: Westview.

Baars, B. J. (2005). Global workspace theory of consciousness: toward a cognitive neuroscience of human experience. *Prog. Brain Res.* 150, 45–53. doi: 10.1016/S0079-6123(05)50004-9

Bence, N. (2014). An experiential account of creativity. In Elliot Samuel Paul and Scott Barry Kaufman, editors, *The Philosophy of Creativity*: New Essays. Oxford University Press.

Blackburn, S. (2017). Creativity and not-so-dumb luck. In S. Elliot Paul & S.B. Kaufman (Eds.), *The philosophy of creativity: New essays* (pp. 147–156). London, UK: Oxford University Press.

Boden, M.A. (2001). Creativity and knowledge. In A. Craft, B. Jeffrey, &M. Leibling (Eds.), *Creativity in education* (pp. 95–102). London, UK: Continuum.

Boden, M. A. (2004). *The creative mind: Myths and mechanisms* (2nd ed.). London, New York: Routledge.

Boden, M.A. (2010). *Creativity and Art: Three Roads to Surprise*, Oxford: Oxford University Press.

Boden, M. A. (2014). Creativity and artificial intelligence: a contradiction in terms. In: Paul, E., Kaufman, S. (Eds.) *The Philosophy of Creativity: New Essays*, pp. 224–246. Oxford University Press, Oxford.

Briskman, Larry. (2009). Creative product and creative process in science and art. In *The concept of creativity in science and art*, Denis Dutton, Micheal Krausz (Eds), Dordrecht: Springer.

Bruno, W. Paul. (2010). *Kant's concept of genius: Its origin and function in the third critique*. London/New York: Continuum.

Butlin P, Long R, Elmoznino E, Bengio Y, Birch J, Constant A, Deane G, Fleming SM, Frith C, Ji X, Kanai R, Klein C, Lindsay G, Michel M, Mudrik L, Peters MAK, Schwitzgebel E, Simon J, VanRulle R. (2023). Consciousness in artificial intelligence: insights from the science of consciousness. [2308.08708] Consciousness in Artificial Intelligence: Insights from the Science of Consciousness

Campbell, Donald. (1960). 'Blind Variation and Selective Retention in Creative Thought as in Other Knowledge Processes.' *Psychological Review*, 67.6, 380–400.

Chen, C. (2025). How DeepSeek became a fortune teller for China's youth: AI-powered BaZi analysis has become the new oracle for a disillusioned generation seeking answers. *MIT Technology Review*.https://www.technologyreview.com/2025/03/03/1112604/deepseek-fortune-teller-china/

Clark, A. (2003). Artificial intelligence and the many faces of reason. In S. Stich & T. Warfield (Eds.), *Blackwell guide to philosophy of mind*. New Jersey: Blackwell.

Clark, A. & Toribio, J. (1994). Doing without representing? *Synthese*, 101 (3), 401–431.

Cropley, David H., Arthur J. Cropley, James C. Kaufman, and Mark A. Runco (Eds.). (2010). The Dark Side of Creativity, New York: Cambridge University Press. doi:10.1017/CBO9780511761225

Csikszentmihalyi, M. (1996). *Creativity: Flow and the psychology of discovery and invention*. New York, NY: Harper Collins.

Csikszentmihalyi, M. (1999). Implications of a system perspective for the study of creativity. In R. J. Sternberg (Eds.), *Handbook of creativity* (pp.313–335). Cambridge, UK: Cambridge University Press.

Dennett, D.C. (1991): *Consciousness Explained*, Cambridge MA: MIT Press/ Bradford Books.

Dreyfus, H. (1992). *What Computers Still Can't Do: A Critique of Artificial Reason*. MIT Press, Cambridge, MA.

Elster, Jon. (2000). *Ulysses Unbound: Studies in Rationality, Precommitment, and Constraints*. Cambridge: Cambridge University Press.

Ericsson, K. Anders, Ralf T. Krampe, and Clemens Tesch-Römer. (1993). "The Role of Deliberate Practice in the Acquisition of Expert Performance", *Psychological Review*, 100(3): 363–406.

Feist, Gregory. J. (1999). The influence of personality on artistic and scientific creativity. In R. J. Sternberg (Ed.), *Handbook of Creativity*. New York: Cambridge University Press.

Finke, Ronald A. (1996). "Imagery, Creativity, and Emergent Structure", *Consciousness and Cognition*, 5(3): 381–393.

Gallagher, S. (2023). *Embodied and Enactive Approaches to Cognition*. Cambridge: Cambridge University Press.

Gardner, H. (1993). *Creating Minds: An Anatomy of Creativity Seen through the Lives of Freud, Einstein, Picasso, Stravinsky, Eliot, Graham, and Gandhi*, New York: BasicBooks.

Gaut, B. (2010). The philosophy of creativity. *Philosophy Compass*, 5(12), 1034-1046.

Gaut, B. (2012). Creativity and rationality. *The Journal of Aesthetics and Art Criticism, 70*, 259–270.

Gaut, B. (2018). "The value of creativity". In *Creativity and Philosophy*. B. Gaut and M. Kieran (Eds), 124–139, New York: Routledge.

Grant, J. (2018). Creativity as an artistic merit. In B. Gaut & M. Kieran (Eds.), In *Creativity and Philosophy*, 333–349. Abingdon, UK: Routledge.

Hamlyn, D.W. (1984). Metaphysics. New York: Cambridge University Press, 4-8.

Hausman, C. R. (1985). Originality as a criterion of creativity. In *Creativity in art, religion, and culture*, H. Mitias (Ed), 26-41. Amsterdam: Rodopi B. V.

Jung, Rex Eugene, Brittany S. Mead, Jessica Carrasco, and Ranee A. Flores. (2013). "The Structure of Creative Cognition in the Human Brain", *Frontiers in Human Neuroscience*, 7.

Kahl, C. H., & Hansen, H. (2015). Simulating creativity from a systems perspective: CRESY. *Journal of Artificial Societies and Social Simulation*, 18.

Kant, I., Pluhar, W. S., & Gregor, M. J. (1987). *Kant's critique of judgement: including the first introduction*. Indianapolis: Hackett Publishing.

Kaufman, James C. (2009). *Creativity 101*, (The Psych 101 Series), New York: Springer Publishing.

Kelley D, Twyman M. (2019). Biasing in an independent core observer model artificial general intelligence cognitive architecture. In: AAAI spring symposia 2019. Stanford University

Kirkpatrick, K. (2023). "Can AI demonstrate creativity?". *Communications of the ACM*, 66(2), 21-23.

Klausen, Søren Harnow. (2010). "The Notion of Creativity Revisited: A Philosophical Perspective on Creativity Research." *Creativity Research Journal*, 22(4):347–360.

Kriegel, U. (2003). Is Intentionality Dependent upon Consciousness? *Philosophical Studies*, 116, 271–307.

Kronfeldner, Maria. (2009). "Creativity Naturalized", *The Philosophical Quarterly*, 59(237).

Kronfeldner, Maria. (2018). "Explaining Creativity." In Berys Gaut, and Matthew Kieran (Eds.), *Creativity and Philosophy*, 213–229. New York: Routledge.

Lovelace, A. A. (1843). Translation of, and notes to, Luigi F. Menabrea's sketch of the analytical engine invented by Charles Babbage. In R. Taylor (Ed.), *Scientific memoirs* (Vol. 3, pp. 691–731). London: Richard and John E. Taylor.

Majeed, R. (2022). The Relationship Between Conscious and Unconscious Intentionality. *Philosophy*, *97*(2), 169–185.

Michael J. A. Howe. (1999). *Genius Explained*. Cambridge, UK: Cambridge University Press.

Mulgan, T. (2018). Moral Imaginativeness, Moral Creativity and Possible Futures. B. N. Gaut (Eds.), In *Creativity and Philosophy*, 351–390. New York: Routledge.

Mumford, M. D., & Gustafson, S. B. (1988). Creativity syndrome: Integration, application, and innovation. *Psychological Bulletin*, 103, 27–43. doi:10.1037/0033-2909.103.1.27

Nagel, T. (1974). What is it like to be a bat? *Philosophical Review*, 83(October), 435-450.

Paul, Elliot Samuel, and Dustin Stokes. (2024). *"Creativity", The Stanford Encyclopedia of Philosophy*. Edward N. Zalta & Uri Nodelman (Eds.). URL = https://plato.stanford.edu/archives/spr2024/entries/creativity/

Paul, Elliot Samuel, and Dustin Stokes. (2021). "Computer Creativity is a Matter of Agency", *Institute of Arts and Ideas News*. Computer creativity is a matter of agency | Dustin Stokes and Elliot Samuel Paul » IAI TV

Paul, E. S., & Stokes, D. (2018). Attributing creativity. In B. Gaut & M. Kieran (Eds.), *Creativity and philosophy* (pp. 193–210). New York: Routledge.

Paul, Elliot. Samuel., & Kaufman, Scott. Barry. (2014). Introducing the philosophy of creativity. *The philosophy of creativity: New essays*, 70(3), 3.

Pease, Alison and Simon Colton. (2011). "On Impact and Evaluation in Computational Creativity: A Discussion of the Turing Test and an Alternative Proposal," in Proceedings of AISB '11: *Computing and Philosophy*, Dimitar Kazakov and George Tsoulas (Eds.), York: Society for the Study of Artificial Intelligence and Simulation of Behaviour, 15–22.

Raichle, M. E. (2015). The brain's default mode network. *Annu. Rev. Neurosci.* **38**, 433–447.

Rhodes, M. (1961). An analysis of creativity. *Phi Delta Kappan*, 42, 305–310.

Rosenberg, A. (2005). *Philosophy of science: A contemporary introduction*. London: Routledge.

Rosenthal, D. M. (1993). Multiple Drafts and Higher-Order Thoughts [Review of *Consciousness Explained*, by D. C. Dennett]. *Philosophy and Phenomenological Research*, *53*(4), 911–918. https://doi.org/10.2307/2108263

Runco, M. A., & Kim, D. (2018). The four Ps of creativity: Person, Product, Process, and Press. In J. Stein (Ed.), *Reference Module in Neuroscience and Biobehavioral Psychology*. Berkeley: Elsevier. https://www.sciencedirect.com/science/article/pii/B9780128093245061939

Sawyer, R. K., & Henriksen, D. (2024). *Explaining creativity: The science of human innovation*. New York: Oxford University Press.

Scott, Dominic. (2011). " Plato, Poetry and Creativity." In *Plato and the Poets*, Destrée and Herrmann (Eds.), 131–154.

Searle, John R. (1980). "Minds, Brains, and Programs", *Behavioral and Brain Sciences*, 3(3): 417–424.

Shirky, C. (2008), "*The power of organizing without organizations*", *Here Comes Everybody*, New York: Penguin Press New York, NY.

Simonton, D. K. (1984). *Genius, creativity, and leadership*. Cambridge, MA: Harvard University Press.

Simonton, D. K. (1999). *Origins of genius: Darwinian perspectives on creativity*. New York: Oxford University Press.

Simonton, D. K. (2010). Creative thought as blind-variation and selective-retention: Combinatorial models of exceptional creativity. *Physics of Life Reviews*, 7(2), 156-179.

Sternberg, R. J., & Lubart, T. (1999). The concept of creativity: Prospects and paradigms. In R. Sternberg (Ed.), *Handbook of Creativity* (pp. 3–15). Cambridge, U.K.: Cambridge University Press.

Sternberg, R. J., & Lubart, T. (1995). *Defying the Crowd: Cultivating Creativity in a Culture of Conformity*, New York, NY: Free Press.

Stokes, Dustin R. (2008). "A Metaphysics of Creativity", in *New Waves in Aesthetics*, Kathleen Stock and Katherine Thomson-Jones (Eds.), New York: Palgrave-Macmillan, 105–124.

Stokes, Dustin R. (2011). "Minimally Creative Thought: Minimally Creative Thought", *Metaphilosophy*, 42(5).

Tomas, V. (1958). Creativity in Art. *The Philosophical Review*, *67*(1), 1–15.

Turing, Alan M. (1950). "Computing Machinery and Intelligence", *Mind*. 59(236): 433–460.

Wallas, Graham. (1926). *The Art of Thought*, London: J. Cape.

Walia, Chetan. (2019). A Dynamic Definition of Creativity. *Creativity Research Journal*, *31*(3), 237–247.

Ward, Thomas B., Steven M. Smith, and Ronald A. Finke. (1999). Creative Cognition. In *Handbook of Creativity*. Robert J. Sternberg (Eds). Cambridge: Cambridge UP, 189–212.

Weisberg, R. (2006). Expertise and reasons in creative thinking. In J. C. Kauffman & J. Baer (Eds.), *Creativity and reason in cognitive development* (pp. 7–42). New York: Cambridge University Press.

Wolfe, C. (2009). *What is posthumanism?* Minneapolis: University of Minnesota Press.

Vandervert, L. R., & Vandervert-Moe, K. J. (2020). Neuroscience: The cerebellum's predominant role in creativity. In M. Runco & S. Pritzker (Eds.), *Encyclopedia of creativity* (3rd ed., Vol. 2, pp. 211–215). Academic Press.

Vygotsky, L. S. (2004). Imagination and creativity in childhood. *Journal of Russian & East European Psychology*, 42,7–97.

Zagzebski, Linda. Trinkaus. (1996). *Virtues of the mind : an inquiry into the nature of virtue and the ethical foundations of knowledge*. New York: Cambridge University Press.